\documentclass{svproc}
\usepackage{url}

\usepackage[noend]{algpseudocode}
\usepackage[T1]{fontenc}
\usepackage{lmodern}
\usepackage{amsmath}
\usepackage{mathtools}
\usepackage[ruled, vlined, linesnumbered]{algorithm2e}

\usepackage{caption}
\usepackage{subcaption}

\begin{document}
\mainmatter              
%

\title{DiabML: AI-assisted diabetes diagnosis method with meta-heuristic-based feature selection} 

\textbf{Proceedings of 14th Turkish Congress of Medical Informatics 16 (18), 19-30}
\titlerunning{Proceedings of 14th Turkish Congress of Medical Informatics 16 (18), 19-30}  
%
\author{Vahideh Hayyolalam
\and \"Oznur \"Ozkasap
}
\authorrunning{V.Hayyolalam \& \"O.\"Ozkasap} 
%
\tocauthor{Vahideh Hayyolalam, \"Oznur \"Ozkasap}
\institute{Department of Computer Engineering, Koç University, İstanbul, Türkiye\\
\email{\{vhayyolalam20; oozkasap\}@ku.edu.tr}\\ 
}

\maketitle              

\begin{abstract}
Diabetes is a chronic disorder identified by the high sugar level in the blood that can cause various different disorders such as kidney failure, heart attack, sightlessness, and stroke. Developments in the healthcare domain by facilitating the early detection of diabetes risk can help not only caregivers but also patients. AIoMT is a recent technology that integrates IoT and machine learning methods to give services for medical purposes, which is a powerful technology for the early detection of diabetes. In this paper, we take advantage of AIoMT and propose a hybrid diabetes risk detection method, DiabML, which uses the BWO algorithm and ML methods. BWO is utilized for feature selection and SMOTE for imbalance handling in the pre-processing procedure. The simulation results prove the superiority of the proposed DiabML method compared to the existing works. DiabML achieves 86.1\% classification accuracy by AdaBoost classifier outperforms the relevant existing methods.

\end{abstract}
\section{Introduction}

    Diabetes is one of the major invasive disorders which rises up in human beings because of the high glucose level in the blood. The long-term disorderliness in glucose levels contributes to severe damage in diabetic patients' blood vessels \cite{Diabet8}. As reported by World Health Organization (WHO), diabetes is known as the seventh fatal disease. Recently, the essential changes in human beings' lifestyles have led to arising various disorders, such as obesity/overweight, hypertension, diabetes, and cancer \cite{Diabet3}. 
    
    Generally, diabetes is caused by some life-risk factors such as high body weight, inappropriate food style, sleeplessness, and chewing tobacco. In the short-term, diabetes can lead to kidney failure, hypoglycemia, and ketoacidosis \cite{Diabet5}. Besides, gangrene can be triggered by diabetic foot diseases. In the long term, diabetes can lead to cardiac diseases, microvascular, and sightlessness \cite{Diabet1}. Diabetes may even appear in children due to the lack of physical activity. Unfortunately, there are not any appropriate medicines for diabetes. However, managing it with suitable measures can lead to a healthy, long, and normal life \cite{Diabet6}. 

    Based on the International Diabetes Federation (IDF) reports, the number of people who die from diabetes is dramatically high. Also, there are numerous people who suffer from diabetes. There is an essential need for technology dedicated to diabetes management and early detection of the risk of getting diabetes \cite{Diabet7}. Adopting emerging technologies, like the Internet of Medical Things (IoMT) \cite{IoMT}, Artificial Intelligence (AI), and Machine Learning (ML), paves the way to identify and predict the risk of getting diabetes in advance \cite{IoMTDiabet}.

    IoMT is the subgroup of the Internet of Things (IoT) technology, which includes connected medical apparatuses for healthcare management and monitoring. IoMT technology integrates automation, AI, and physical sensors, which facilitates the self, clinical, and remote monitoring of the patients. The integration of the IoT, medical devices, and ML models, along with data analytics procedures for advanced and complicated healthcare services, contributed to a novel concept called Artificial Intelligence of Medical Things (AIoMT) \cite{AIoMT}. Using AIoMT, patients can easily connect to health centers/specialists via medical apparatuses and remotely access, collect, transmit, and process health data through a secure network. AIoMT helps with diminishing inessential hospitalization and, thereby, the corresponding healthcare costs by opening the doors for wireless monitoring of health factors.

    The integration of medical devices and AI, especially ML, has extremely boosted healthcare services. This transformation from traditional healthcare services/monitoring to automated, connected AIoMT-based health services forms a revolution in the medical field. Owing to these technologies, people with reduced cost and time are able to track their health condition \cite{vh}.
    
    Accordingly, adopting the integration of AIoMT can pave the way for the early detection of diabetes or the risk of getting it, which helps to reduce the detrimental effects. This research aims to improve the early detection of diabetes by reducing the number of features using a meta-heuristic algorithm, namely Black Widow Optimization (BWO) \cite{bwo}. This work includes comparative research on different ML classifiers applied to a publicly available dataset. The contributions of this study are:
    
\begin{itemize}
    \item  We propose a hybrid method, DiabML, for the early detection of diabetes, which utilizes a meta-heuristic algorithm and ML classifiers. In other words, a hybrid meta-heuristic and ML approach is proposed to detect the risk of diabetes in advance. The designed strategy adopts the BWO algorithm to reduce the number of features and ML classifiers to predict diabetes. 

    \item A publicly available dataset, named Diabetes Health Indicators Dataset, accessible in Kaggle, is adopted. Before conducting any processing on data, we normalize them between 0 and 1 since their ranges and units were different. Normalizing the data helps with placing them in one range and facilitating data processing. To this end, we have utilized the MinMax scalar technique.
    
    \item  Since the adopted dataset is an imbalanced dataset, we utilize Synthetic Minority Oversampling Techniques (SMOTE) to cope with the imbalance data problem.
    
    \item   DiabML method achieves 86.1\% classification accuracy by AdaBoost classifier, which outperforms the relevant existing works. The experiments prove the impact of handling data imbalance issue in classification.
    
\end{itemize}

    The remainder of the paper is as follows. Related literature is discussed in section \ref{sec.rel}. The description of system model is provided in section \ref{sec.model}. Then, section \ref{sec.data} includes the description of the adopted dataset and the details of the proposed DiabML methodology. The outcomes of the research are brought into sight in section \ref{sec.result}. Eventually, the conclusion of this study and future work are discussed in section \ref{sec.conclusion}.

\section{Related Literature} \label{sec.rel}

    There exist several research works attempting to diagnose diabetes in advance. In this section, the studies that focus on predicting the risk of getting diabetes using ML classifiers are reviewed.

    The authors in \cite{Diabet2} proposed an automated diabetes detection system using ML models. The authors have utilized five ML classifiers, including Logistic Regression (LR), Decision Tree (DT), Random Forest (RF), K-Nearest Neighbor (KNN), and Gaussian Naive Bayes (NB). In order to diminish the dataset dimensionality, they have used Principal Component Analysis (PCA). Also, to handle the imbalance problem, they have utilized SMOTE. They compared the results according to the accuracy and some other metrics. They claimed that Random Forest, with 82.26\% accuracy, defeated other classifiers.

    As another research work, authors in \cite{Diabet4} have tried to predict the risk of diabetes by ML classifiers. They adopted KNN, LR, DT, RF, and NB classifiers and a publicly available dataset. They concluded that the DT and RF, respectively, by 84.78\% and 84.89\% accuracy, outperform other classifiers. They claimed that DT and RF are the two best classifiers for predicting the risk of diabetes.

    Moreover, researchers in \cite{Diabet1} have used Kernel entropy component analysis to decrease the dataset dimensionality. Also, the adopted NB, LR, linear discriminant analysis, KNN, support vector classifier (SVM), DT, extreme gradient boosting, RF, and kernel entropy component analysis. They have three phases in their proposed model, including feature extraction, classification, and diabetes risk prediction.

    Furthermore, researchers in \cite{Diabet12} have utilized NB, LR, SVM, Bayesian Network (BN), Artificial Neural Network (ANN), KNN, RF, Logistic model tree, reduced error pruning, rotation forest, AdaBoost, Stochastic Gradient Descent (SGD) to predict the risk of diabetes. They evaluated the classifiers with various metrics such as Precision, F1-score, recall, and accuracy. They have validated their evaluations by 10-fold cross-validation. Also, they have used SMOTE for the imbalance problem handling.

    Authors in \cite{Diabet9} aim to explore various ML classifiers and their performance to predict the risk of diabetes. They have adopted radial basis functions, KNN, NB, extra trees, Multi-layer perception (MLP), and DT to classify the diabetic data. The classifiers are evaluated by some metrics, such as accuracy, Kappa score, recall, precision, F1-score, root-mean-square error (RMSE), and mean square error (MSE). They have used the Prima Indian diabetes dataset and concluded that MLP outperforms the other classifiers.

    With regard to the reviewed research works above, several studies have been conducted to predict the risk of diabetes via ML, especially by classification algorithms. However, there is still an essential demand for development to obtain more better and accurate outcomes. In this regard, this research designs a novel hybrid strategy, named DiabML, utilizing BWO and ML classifiers to predict the risk of diabetes in advance.

\section{System Model} \label{sec.model}
    The 
    system model includes four layers, encompassing Things, Edge, Fog, and Cloud layers. The Things layer includes different sensors and medical devices which are responsible for data collection. The edge layer comprises small and resource constraint end devices such as smartphones, smart bands, smart watches, and tablets. The fog layer includes devices such as laptops, desktop PCs, and workstations, which are more powerful than edge layer devices. The cloud layer, including cloud data centers, has the capability of processing and storing big data.
    
\begin{figure}[!ht]
    \centering
    \includegraphics[width=1\linewidth]{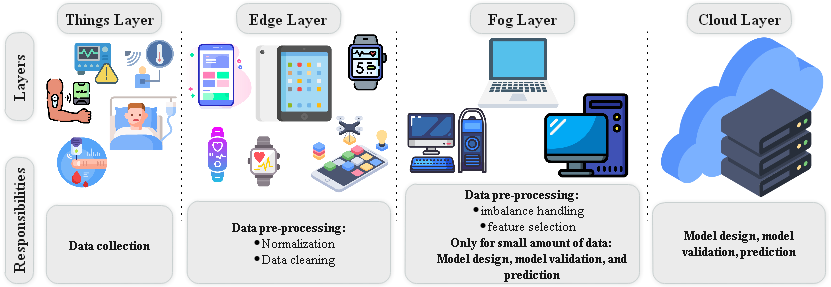}
    \caption{The system layers and phases}
    \label{fig.sys}
\end{figure}

    The general illustration of the system layers and phases is given in Figure \ref{fig.sys}. We defined the AIoMT with these four layers, where the edge, fog, and cloud layers are equipped with ML methods. Since edge and fog layers are close to the end devices, the processes can be conducted on them with less communication cost, less network traffic, less bandwidth usage, less power consumption. 
    
    Our assumption is that the patients are provided with different relevant sensors and IoT devices to collect the needed data. In this architecture, the edge layer is in charge of performing normalization and data cleaning the collected data, then, transmitting them to suitable fog devices. The fog layer is responsible for conducting the  rest of pre-processing procedure, including imbalance handling and feature selection. In case the processing power is enough for applying the classification algorithms on the data, they will be classified in the fog layer. The cloud layer is responsible for performing the processes needed for more powerful systems and permanent results/data storage.

 \vspace{-10pt}

\begin{figure}[!ht]
    \centering
    \includegraphics[width=1\linewidth]{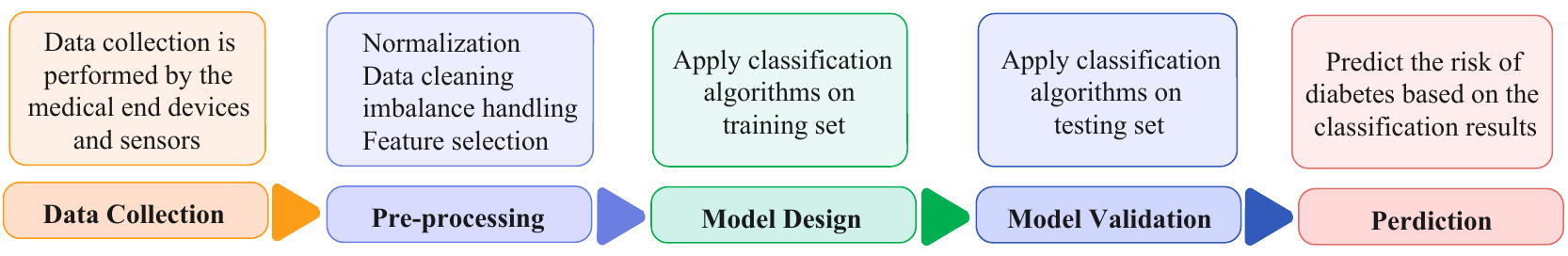}
    \caption{The phases of the proposed methodology DiabML} \vspace{-20pt}
    \label{fig.methflow}
\end{figure}

    A hybrid machine learning strategy, DiabML, for predicting and early detection of diabetes has been proposed in this research. As depicted in Figure \ref{fig.methflow}, the proposed model includes five phases, namely data collection, pre-processing, model design, model validation, and prediction phases. The sensors and other IoT end devices are in charge of performing the data collection phase. The pre-processing phase is responsible for preparing the data for the classification phase. To this end, in this phase, data are normalized between 0 and 1 to scale all data in a shared range. The imbalance handling and dimensionality reduction processes are also accomplished in this phase. The dataset is split into two parts, including the training and testing parts. In the model design phase, various classification methods are applied to the training set. Then, in the model validation phase, the trained models are tested using the test set. Eventually, according to the outcomes, the system would determine whether the associated patient has a risk of diabetes or not. \vspace{-3pt}
    
\section{Data and Methodology} \label{sec.data}
\begin{figure}[!ht]
    \centering
    \includegraphics[width=0.7\linewidth]{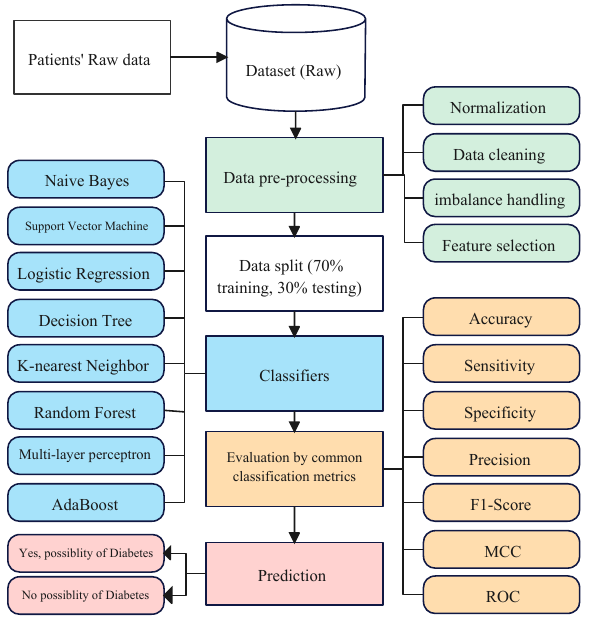}
    \caption{The flowchart of DiabML} \vspace{-10pt}
    \label{fig.flow}
\end{figure}

    This section discusses the used dataset and the proposed smart methodology for diagnosing diabetes in advance. First, the dataset is introduced, and then the flow of the diabetes diagnosis process is thoroughly addressed. Figure \ref{fig.flow} visualizes the flow of the process vividly. The proposed methodology includes five phases, including data acquisition (collecting data from patients), data pre-processing (normalization, data cleaning, imbalance handling, feature selection), classification, evaluation, and prediction.
    
\subsection{Data description}

Diabetes Health Indicators Dataset \cite{dataset} 
is utilized to evaluate the proposed DiabML methodology. This dataset includes $253,680$ data points with 21 features that are collected annually by the CDC via the Behavioral Risk Factor Surveillance System (BRFSS). The adopted dataset is a labeled dataset and is publicly available on Kaggle.

\subsection{Data pre-processing}
    Aiming to classify and predict the risk of diabetes, the raw data should undergo the pre-processing procedure. To this end, first, we normalize the data by MinMax scalar technique to put all the data in a common range. In the next step, for data cleaning, we have adopted Sci-kit learn functions to get rid of dummy data. Afterward, we utilized SMOTE technique to cope with the imbalance problem of data. Then, we used the BWO algorithm for feature selection and reducing the data dimensionality. \vspace{-3pt}

\subsection{Feature selection using BWO}
    Feature selection problem in ML methods is one of the challenging issues for reducing the data volume. The aim of feature selection is to diminish the number of features and select an adequate number of features that lead to the best possible performance. The feature selection is an NP-hard issue since there can be various combinations of features with different performances. Meta-heuristic algorithms are one of the popular and efficient methods for solving NP-hard problems \cite{symb}.

    In this study, the BWO algorithm is adopted to attain an optimal solution for the feature selection issue. The pseudo-code of adopted BWO is illustrated in Algorithm 1
    . The BWO is an evolutionary algorithm imitating the strange habits of black widow spiders. Like other evolutionary algorithms, BWO is initiated by a set of random potential solutions. The procreate phase is applied to the initial set and produces offspring for each pair. The number of offspring will reduce by the cannibalism phase. The existing solutions are modified by the mutation phase and contribute to the new solutions. At the last state, the number of existing solutions must be reduced to reach the primary set size. To this end, the algorithm omits the worst solutions. This algorithm is an iteration algorithm, so all the steps will repeat till the stopping criteria is met. Eventually, the best resolution will be introduced as the result for the associated problem.

    Each potential solution is implemented as a vector, the length of which is equal to the number of selected features ($n$). Each item of the array includes an integer value between 1 and 21 since the number of features in the dataset is 21. Each item represents the index of the chosen feature. In order to determine the number of selected features, we have accomplished some experiments and finally concluded that selecting nine features out of 21 contributes to better results in terms of classification accuracy.   \vspace{-15pt}

    \begin{algorithm}[!ht] \label{alg.BWO}
    \SetAlgoLined
    \textbf{Input} {The dataset including all features}\\
    \textbf{Output} A subset of chosen features\\
         produce random initial solutions \\
     Select better solutions and place them in \textit{p1}\\
     \While{Stop criteria is not met}{
        \For{procreate and cannibalism steps}{
            randomly select a pair from $p1$\\
            generate offspring (solutions)\\
            keep the better solution and remove the other one\\
            apply the cannibalism and remove some offspring\\
            preserve the remaining in \textit{p2}\\
        }
        \For{mutation step}{
            choose a vector from $p1$\\
            update the chosen vector\\
            preserve it in $p3$\\
        }
       append $p2$ and $p3$ to the primary set\\
       select $n$ better solutions and remove the others\\
       \textbf{return} the best solution as a solution of this generation
     }
     \textbf{return} the best result of all iterations as the final solution (A set of selected features)
     \caption{BWO Algorithm}
    \end{algorithm} \vspace{-5pt}

\subsection{Machine learning classifiers}

    This part of the paper is dedicated to introducing the adopted ML classifiers in this study.
        
    \textbf{Naive Bayes} classifier utilizes the Bayes theorem. This classifier is based on the probability and calculates the posterior probability of the class, which indicates the probability of occurrence of that class. \textbf{Logistic Regression} classifier uses the equation \ref{equ.lr} to find the probability of classifying an input X as the member of class 1.
    
    \begin{equation} \label{equ.lr}
        P(X) = \frac{exp(\beta_0 + \beta_1 X)}{1 + exp(\beta_0 + \beta_1 X)}
    \end{equation}
    
    Here $\beta_0$ represents the bias, and $\beta_1$ indicates the weight, which should be multiplied by input X.
       
    \textbf{Decision Tree} classifier includes interlinks among both external and internal nodes, which are used for decision-making. \textbf{K-Nearest Neighbors}classifier tries to find the distances among an entity and all the data points in the dataset, choosing the determined number of data points (K) nearest to the associated entity, then deciding on the most frequent class label. \textbf{Random Forest} classifier combines various decision trees by utilizing the bagging concept. This combination can somehow improve prediction accuracy \cite{Diabet6}. \textbf{SVM} classifier generates a hyperplane in order to conduct the classification. Thus, all of the data points corresponding to one class would place on one side of the hyperplane and the rest on the other side. The aim is to optimize the maximum distance between the two classes. \textbf{Multilayer perceptron} classifier is one of the popular neural network models in the connected healthcare field, which take advantage of using different layers. \textbf{AdaBoost} classifier is an adaptive boosting algorithm that utilizes the ensemble strategy for increasing the efficiency of weak learners.
    
\subsection{Evaluation metrics}
    
    In this section, performance metrics used for the evaluation and comparison of the proposed DiabML methodology are described. The adopted evaluation metrics are usually defined and calculated according to FP, TP, FN, and TN, which are described below.

\begin{itemize}
    \item \textbf{FP} stands for False positive, defined as the wrong anticipation ratio of the positives.
    \item \textbf{TP}  the true classification ratio for the positive class.
    \item \textbf{FN} stands for False negative, defined as the wrong prediction ratio of the negatives.
    \item \textbf{TN} stands for True Negative, defined as the true classification ratio of the negative class \cite{Diabet10}.
\end{itemize}

The utilized metrics are listed and described as follows.

\textbf{ROC} (Receiver Operating Characteristic) \textbf{Curve} is a graph that represents the false positive ratio versus true positive ratio (FPR vs. TPR defined by equations \ref{eq2} and \ref{eq1}).

\begin{equation}
FPR=\frac{FP}{(FP+TN)}
\label{eq2}
\end{equation}

\begin{equation}
TPR=\frac{TP}{(TP+FN)}
\label{eq1}
\end{equation}

The area under the ROC curve shows AUC (the area under the ROC Curve), which displays the cumulative measurement of all possible classification thresholds.

 \textbf{Accuracy} is a primary measure to compute the performance of classification problems by a mathematical formula  defined in equation \ref{eq3}. 

\begin{equation}
Accuracy=\frac{(TP+TN)}{(TP+FN+TN+FP)}
\label{eq3}
\end{equation}

\textbf{Sensitivity}: Recall (Sensitivity) is defined as equation \ref{eq4}, and identifies any patient having Diabetes.  

\begin{equation}
Sensitivity=\frac{TP}{(TP+FN)}
\label{eq4}
\end{equation}

\textbf{Specificity:} is defined as equation \ref{eq5} which identifies any non-Diabetic patient.  
\begin{equation}
Specificity=\frac{TN}{(TN+FP)}
\label{eq5}
\end{equation}

\textbf{Precision}: There is another important evaluation metric named Precision that computes how the model performs precisely. Precision can be defined as equation \ref{eq6}.

\begin{equation}
Precision=\frac{TP}{(TP+FP)}
\label{eq6}
\end{equation}

\textbf{ F1-score:} Discovering an equilibrium and balance between precision and sensitivity is the aim of the F1-score, which can be defined by equation \ref{eq7}.

\begin{equation}
F1-score =\frac{(2TP)}{(2TP+FN+FP)}
\label{eq7}
\end{equation}

\textbf{MCC:} Finally, MCC is the abbreviate of Mathew Correlation Coefficient and is defined as equation \ref{eq9}.

\begin{equation}
  MCC=\frac{(TP\times TN) - (FP\times FN)}{\sqrt{(TP+FP)(TP+FN)(TN+FP)(TN+FN)}}
 \label{eq9}
\end{equation}

\section{Experiments and Results} \label{sec.result}

This section discusses the experiments and the obtained outcomes. To evaluate the proposed DiabML method, we execute the implementation with/without feature selection and with/without imbalance handling. 
    
In the first part of the demonstration of the results, we have compared the obtained model accuracy of our proposed DiabML method with two existing works, PCAML \cite{Diabet2}, and Vanilla \cite{Diabet4}, which adopted the same dataset. Table \ref{tab:my_label} illustrates the accuracy comparison between the proposed method and existing works, as mentioned earlier. Some of the classifiers we adopted have not been used in the existing works, which are marked as N/A (that is, Not Applied) in the table. Considering these results, it can be concluded that our proposed method, almost in all classifiers, outperforms the other existing works. \vspace{-3pt}

  \begin{table}[h!]
    \centering
     \caption{Accuracy comparison with existing works} 
    \begin{tabular}{|c|c|c|c|}
    \hline
         Algorithm & Proposed DiabML & PCAML \cite{Diabet2} & Vanilla \cite{Diabet4} \\
    \hline
        NB & \textbf{81.23} & 70.56 & 79.02\\ 
    \hline
        SVM & \textbf{85.95} & N/A & N/A\\ 
    \hline
        LR & \textbf{86.01} & 72.64 & 72.49\\ 
    \hline
        DT & \textbf{86.08} & 81.02 & 84.78\\ 
    \hline
        KNN & \textbf{84.26} & 80.55 & 74.27\\ 
    \hline
        RF & \textbf{85.12} & 82.26 & 84.89\\
    \hline
        MLP & \textbf{86.01} & N/A & N/A\\ 
    \hline
        AdaBoost & \textbf{86.1} & N/A & N/A\\
    \hline
    \end{tabular}
   
    \label{tab:my_label}
\end{table}

\begin{figure*}[h!]
	\centering
\subfloat[Sensitivity comparison 
]{
		{\includegraphics[width=0.45\linewidth]{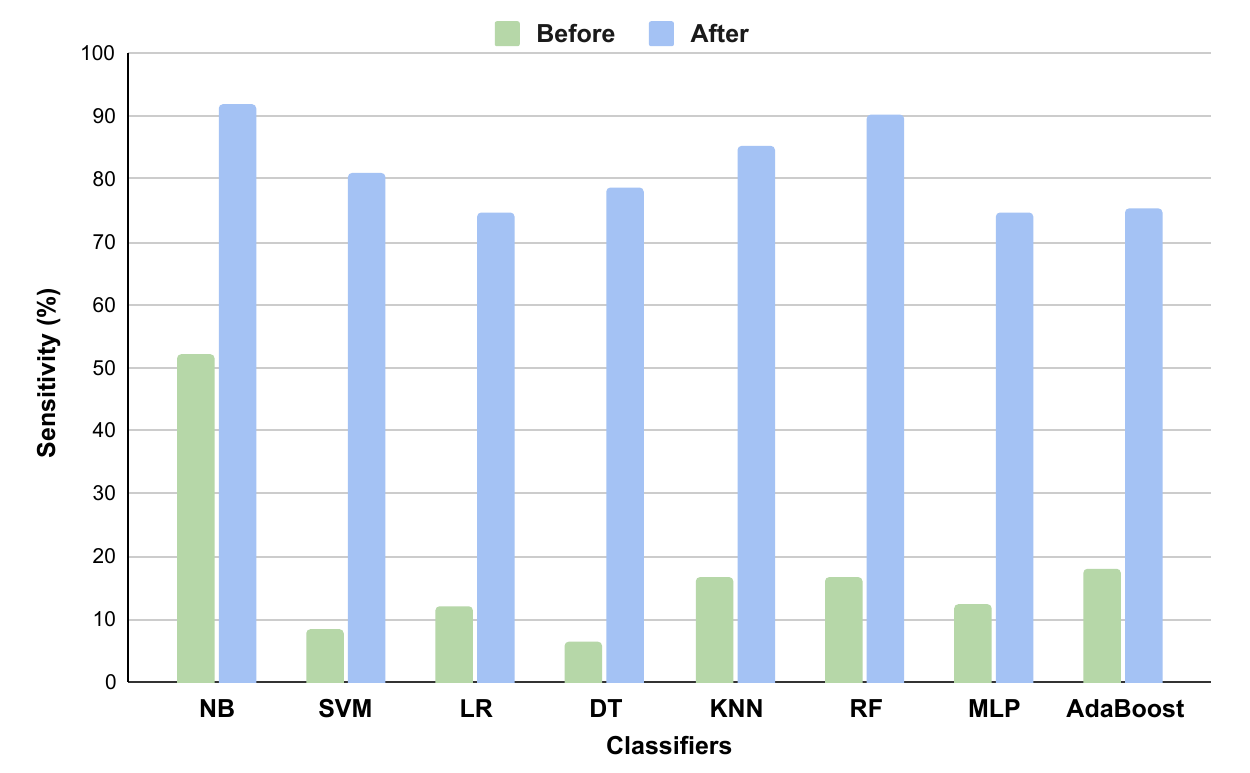}}
		\label{fig.sen}
	}
	\hfill
	\subfloat[Precision comparison 
	]{
		{\includegraphics[width=0.45\linewidth]{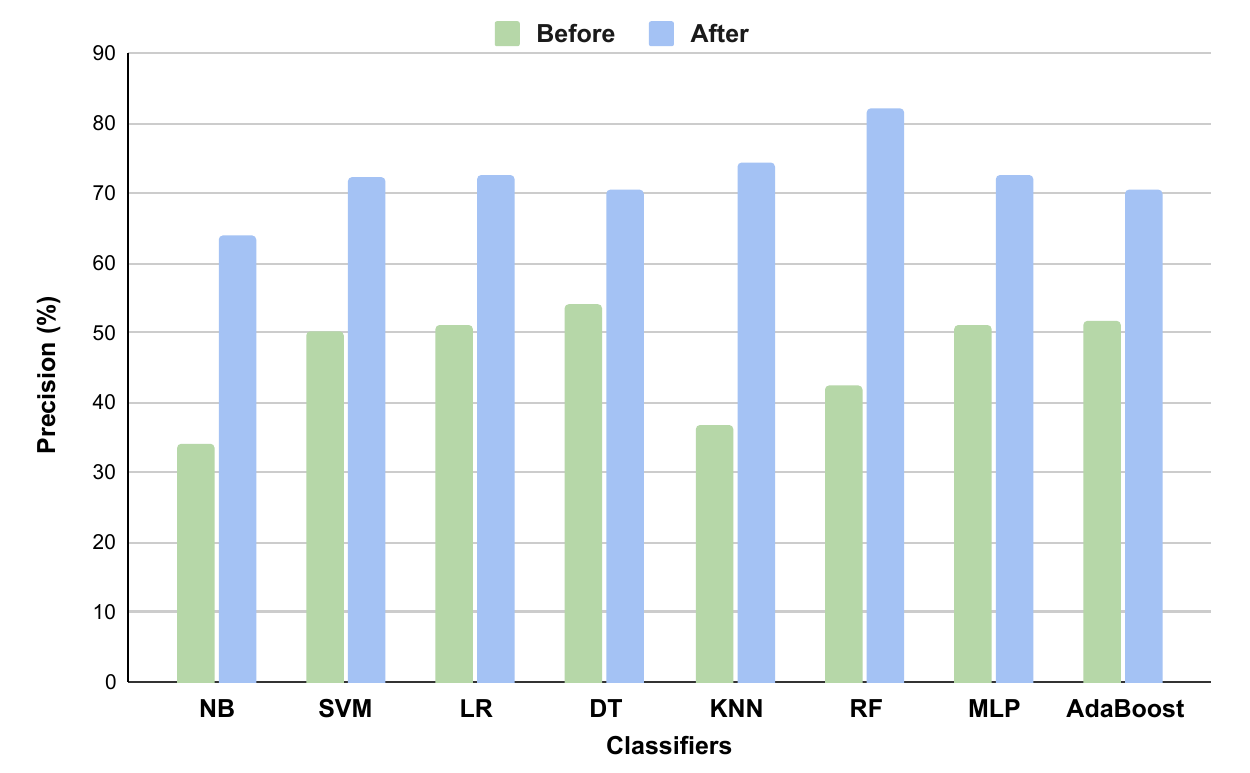}}
		\label{fig.prec}
		}
			\hfill
	
		\caption{Classifiers Performance before and after imbalance handling} \vspace{-15pt}
	\label{fig.acc.preSens}
\end{figure*}

\begin{figure*}[h!]
	\centering
\subfloat[F1-Score comparison 
]{
		{\includegraphics[width=0.45\linewidth]{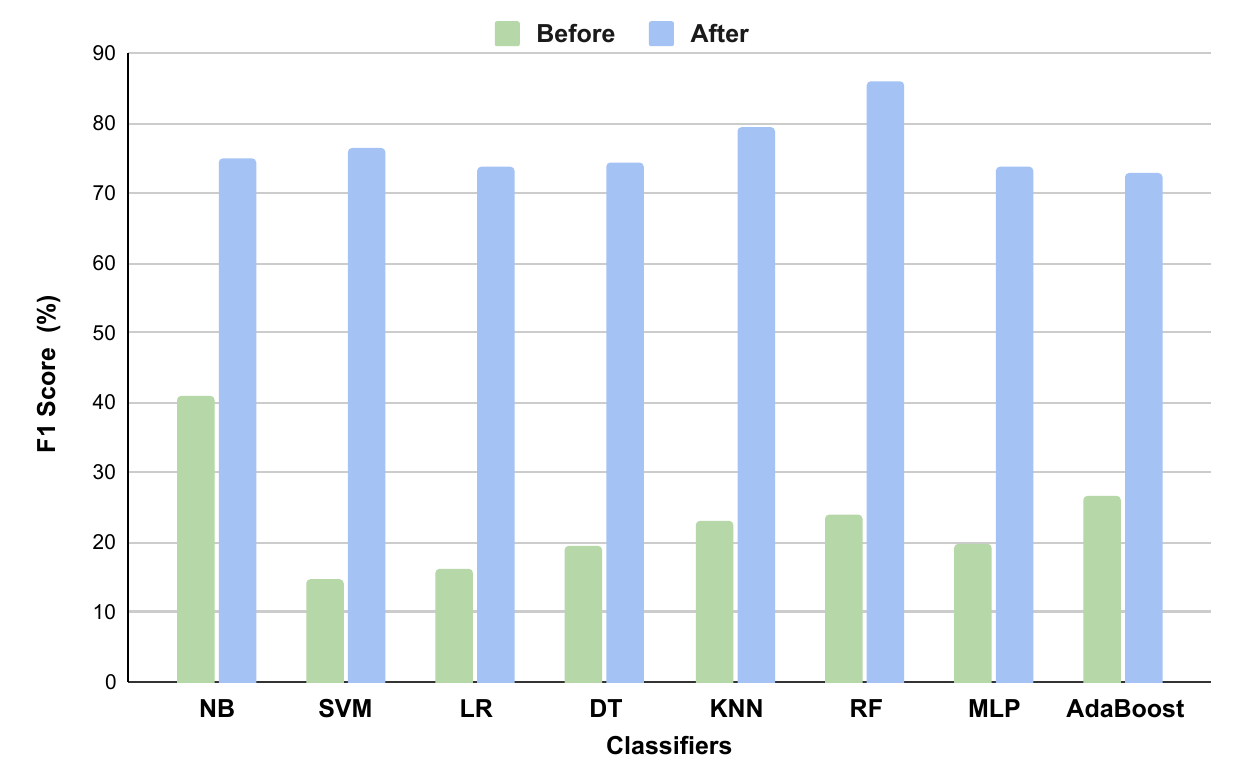}}
		\label{fig.f1}
	}
	\hfill
	\subfloat[MCC comparison 
	]{
		{\includegraphics[width=0.45\linewidth]{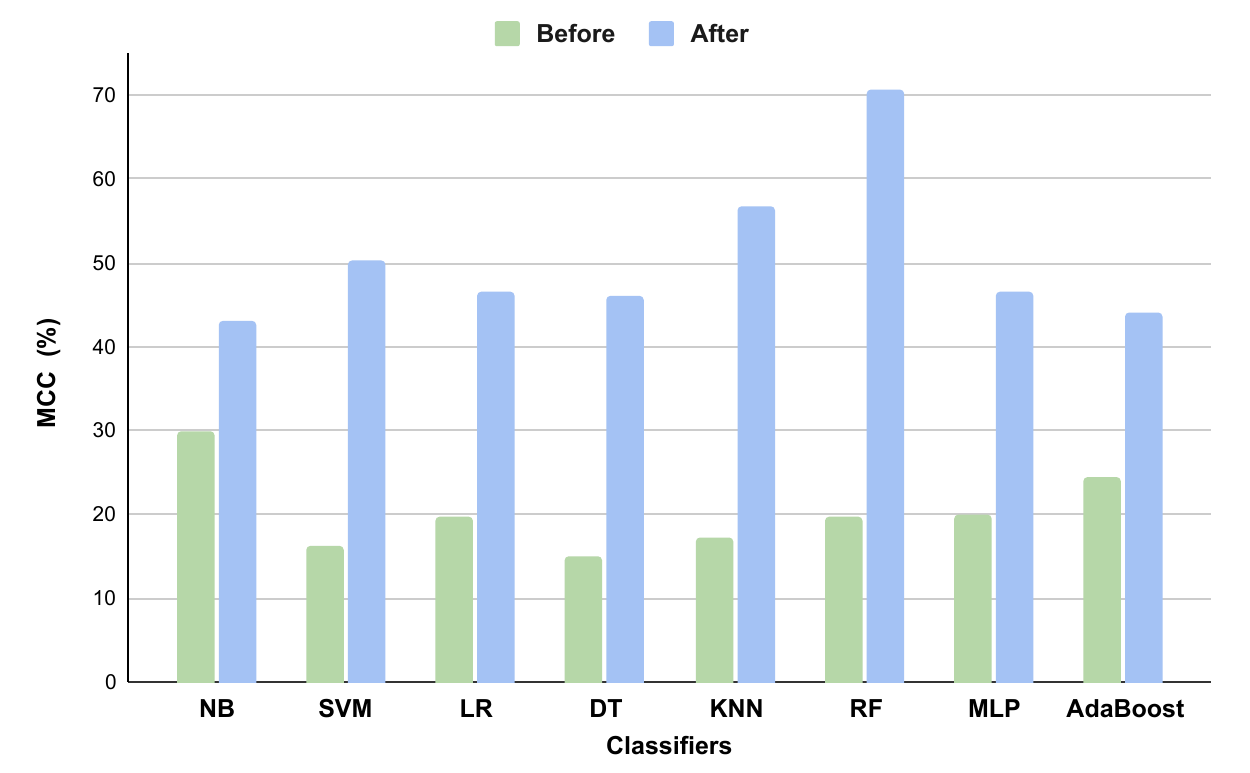}}
		\label{fig.mcc}
		}
			\hfill
	
		\caption{Classifiers Performance before and after imbalance handling} \vspace{-10pt}
	\label{fig.f1MCC}
\end{figure*}

    In the second part, the implementation is tested with/without feature selection/imbalance handling. Figure\ref{fig.acc.preSens} depicts sensitivity and precision before and after imbalance handling in subfigures \ref{fig.sen} and \ref{fig.prec} respectively. Similarly, Figure \ref{fig.f1MCC} compares F1-score and MCC of different classifiers before and after handling the data imbalance problem in Figures \ref{fig.f1} and \ref{fig.mcc}, respectively. As can be seen from the figures, imbalance handling can significantly increase sensitivity, precision, F1-score, and MCC.  
    
 \vspace{-3pt}

\begin{figure*}[h!]
	\centering
\subfloat[Accuracy comparison before and after feature selection
]{
		{\includegraphics[width=0.45\linewidth]{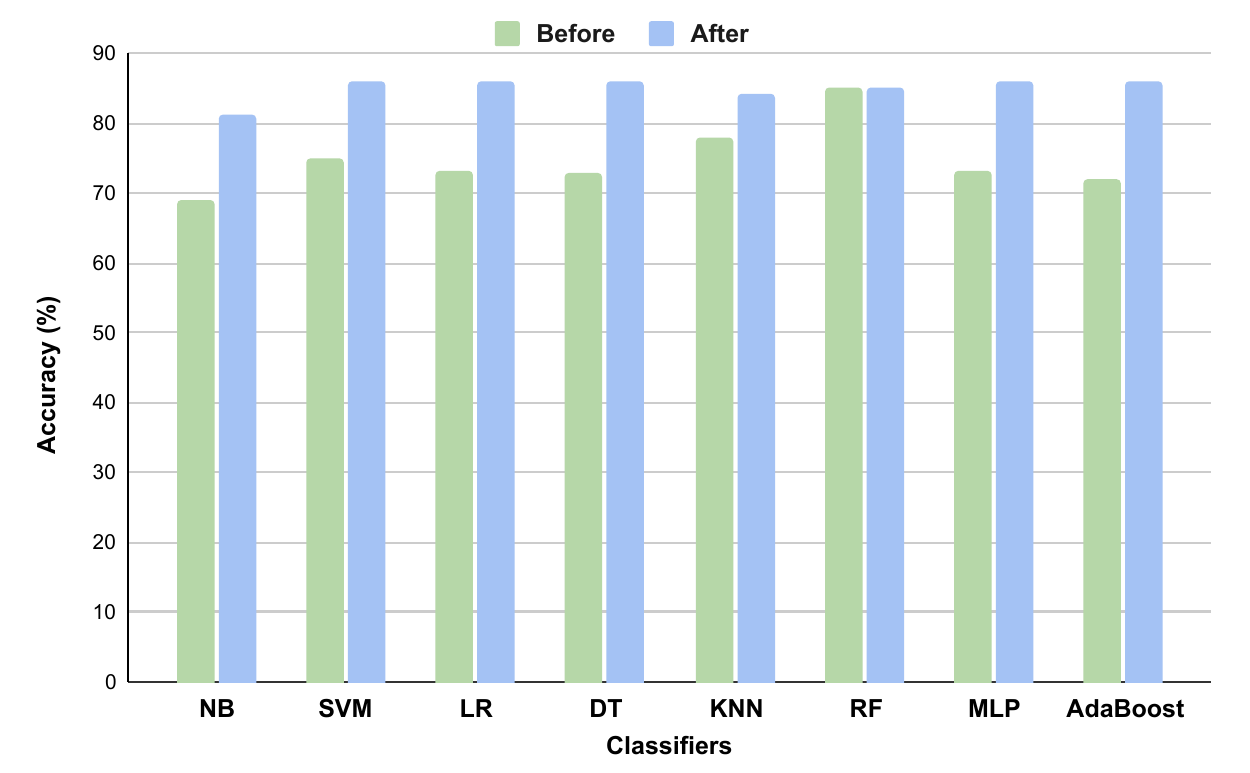}}
		\label{fig.acc}
	}
	\hfill
	\subfloat[ROC curve evaluation
	]{
		{\includegraphics[width=0.45\linewidth]{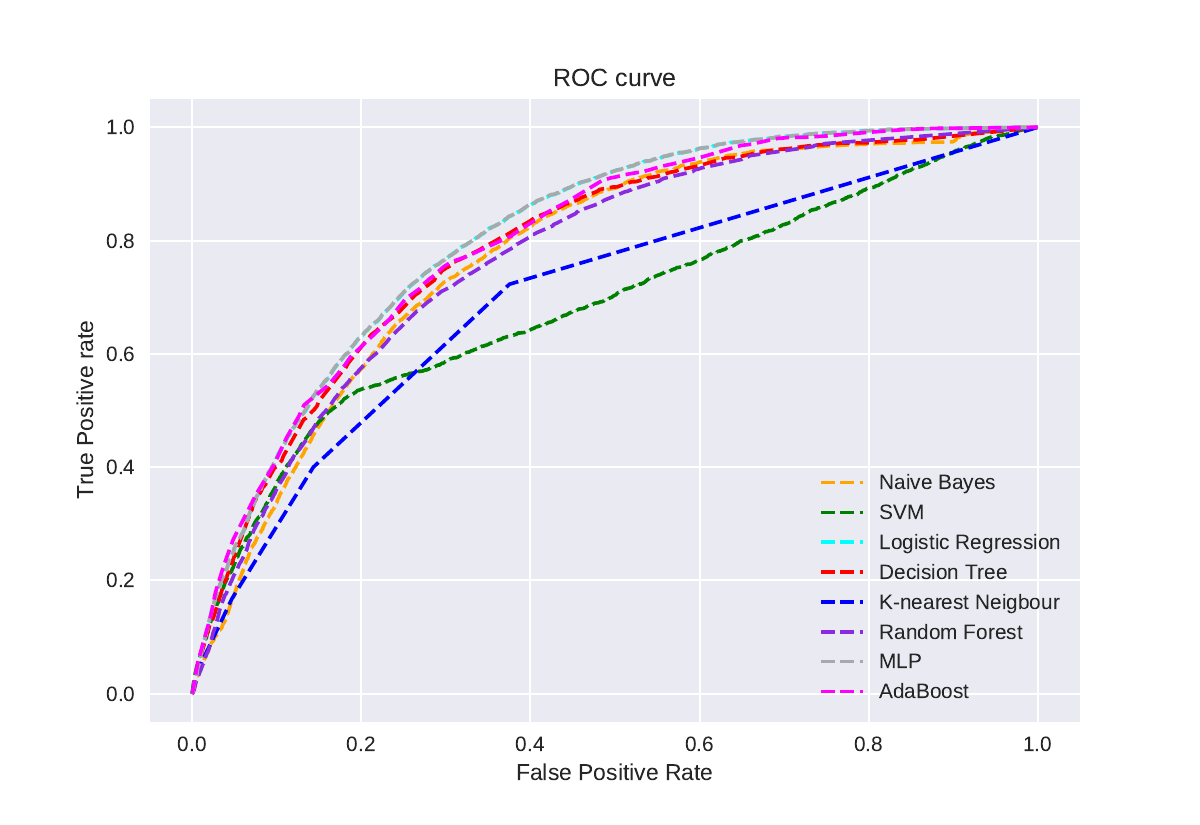}}
		\label{fig.roc}
		}
			\hfill
	
		\caption{Classifiers Performance} \vspace{-15pt}
	\label{fig.acroc}
\end{figure*}

 The comparison of model accuracy with/without feature selection is shown in figure \ref{fig.acc}. As can be seen, all of the classifiers perform better after applying the proposed feature selection strategy. Moreover, figure \ref{fig.roc} shows the ROC curve, which illustrates the trade-off between sensitivity and specificity. The curves within reach of the top-left corner belong to classifiers with better performance. Oppositely, the curves far away from the top-left belong to the classifiers that perform worse. 

\section{Conclusion} \label{sec.conclusion}
    With regard to the severe issues and disorders that can be raised via diabetes, early detection and helping people not to suffer from diabetes is so vital. In this regard, newly emerged technologies like IoMT and AIoMT are the perfect candidates. AIoMT integrated IoT and machine learning to facilitate medical services everywhere for everyone. In this paper, we designed an efficient method, named DiabML, for early diabetes risk detection, which employs a meta-heuristic algorithm named BWO for feature selection and SMOTE for handling the imbalance problem of the adopted dataset. The AIoMT architecture includes four layers, encompassing Things, edge, fog, and cloud data center. The experimental results of the DiabML method are compared to existing works that adopted the same dataset. Also, the results of the DiabML method are compared with the vanilla setups (without feature selection/ imbalance handling). The outcomes demonstrate that the designed DiabML method outperforms the existing works in model accuracy. Also, the results show the impact of feature selection and imbalance handling in terms of accuracy, precision, sensitivity, specificity, F1-Score, and  MCC. Moreover, among the adopted classifiers, AdaBoost, with 86.1\% accuracy, outperforms the other classifiers. In the future, we can conduct learning the data locally instead of collecting data from different clients and producing a big dataset. To this end, federated learning can be utilized to train the data locally and only share the model parameters with other collaborators.

\section*{Acknowledgement}
\noindent This work was supported in part by TUBITAK 2247-A Award 121C338.
%
%
\begin{thebibliography}{18}
\bibitem {Diabet8}
Howlader, K. C., Satu, M., Awal, M., Islam, M., Islam, S. M. S., Quinn, J. M., \& Moni, M. A. (2022). Machine learning models for classification and identification of significant attributes to detect type 2 diabetes. Health information science and systems, 10(1), 1-13.

\bibitem {Diabet3}
Krishnamoorthi, R., Joshi, S., Almarzouki, H. Z., Shukla, P. K., Rizwan, A., Kalpana, C., \& Tiwari, B. (2022). A novel diabetes healthcare disease prediction framework using machine learning techniques. Journal of Healthcare Engineering, 2022.

\bibitem {Diabet5}
Selvakumar, S., Kannan, K. S., \& GothaiNachiyar, S. (2017). Prediction of diabetes diagnosis using classification based data mining techniques. International Journal of Statistics and Systems, 12(2), 183-188.

\bibitem {Diabet1}
Thotad, P. N., Bharamagoudar, G. R., \& Anami, B. S. (2023). Diabetes disease detection and classification on Indian demographic and health survey data using machine learning methods. Diabetes \& Metabolic Syndrome: Clinical Research \& Reviews, 17(1), 102690.

\bibitem {Diabet6}
Chang, V., Bailey, J., Xu, Q. A., \& Sun, Z. (2022). Pima Indians diabetes mellitus classification based on machine learning (ML) algorithms. Neural Computing and Applications, 1-17.

\bibitem {Diabet7}
Maniruzzaman, M., Rahman, M., Ahammed, B., \& Abedin, M. (2020). Classification and prediction of diabetes disease using machine learning paradigm. Health information science and systems, 8(1), 1-14.

\bibitem {IoMT}
Manickam, P., Mariappan, S. A., Murugesan, S. M., Hansda, S., Kaushik, A., Shinde, R., \& Thipperudraswamy, S. P. (2022). Artificial intelligence (AI) and internet of medical things (IoMT) assisted biomedical systems for intelligent healthcare. Biosensors, 12(8), 562.

\bibitem {IoMTDiabet}
Kukkar, A., Gupta, D., Beram, S. M., Soni, M., Singh, N. K., Sharma, A., ... \& Rizwan, A. (2022). Optimizing deep learning model parameters using socially implemented IoMT systems for diabetic retinopathy classification problem. IEEE Transactions on Computational Social Systems.

\bibitem {AIoMT}
Pandya, S., Ghayvat, H., Reddy, P. K., Gadekallu, T. R., Khan, M. A., \& Kumar, N. (2022). COUNTERSAVIOR: AIoMT and IIoT enabled Adaptive Virus Outbreak Discovery Framework for Healthcare Informatics. IEEE Internet of Things Journal.

\bibitem {vh}
Hayyolalam, V., Aloqaily, M., Özkasap, Ö., \& Guizani, M. (2021). Edge-assisted solutions for IoT-based connected healthcare systems: a literature review. IEEE Internet of Things Journal, 9(12), 9419-9443.

\bibitem {bwo}
Hayyolalam, V., \& Kazem, A. A. P. (2020). Black widow optimization algorithm: a novel meta-heuristic approach for solving engineering optimization problems. Engineering Applications of Artificial Intelligence, 87, 103249.

\bibitem {Diabet2}
Chang, V., Ganatra, M. A., Hall, K., Golightly, L., \& Xu, Q. A. (2022). An assessment of machine learning models and algorithms for early prediction and diagnosis of diabetes using health indicators. Healthcare Analytics, 2, 100118.

\bibitem {Diabet4}
Özsezer, G., \& Mermer, G. (2022). Diabetes Risk Prediction with Machine Learning Models. Artificial Intelligence Theory and Applications, 2(2), 1-9.

\bibitem {Diabet12}
Dritsas, E., \& Trigka, M. (2022). Data-driven machine-learning methods for diabetes risk prediction. Sensors, 22(14), 5304.

\bibitem {Diabet9}
Theerthagiri, P., Ruby, A. U., \& Vidya, J. (2023). Diagnosis and Classification of the Diabetes Using Machine Learning Algorithms. SN Computer Science, 4(1), 1-10.

\bibitem {symb}
Hayyolalam, V., Pourhaji Kazem, A.A.: QoS-aware optimization of cloud service composition using symbiotic organisms search algorithm. J. Intell. Proced. Electr. Technol. 8(32), 29–38 (2017)

\bibitem {Diabet10}
Khandakar, A., Chowdhury, M. E., Reaz, M. B. I., Ali, S. H. M., Kiranyaz, S., Rahman, T., \& Hasan, A. (2022). A Novel Machine Learning Approach for Severity Classification of Diabetic Foot Complications Using Thermogram Images. Sensors, 22(11), 4249.


\bibitem {dataset}
    Kaggle publicly available dataset:\\ \Url{https://www.kaggle.com/datasets/alexteboul/diabetes-health-indicators-dataset}


\end{document}